\documentclass[sigconf,nonacm]{acmart}

\AtBeginDocument{%
  \providecommand\BibTeX{{%
    \normalfont B\kern-0.5em{\scshape i\kern-0.25em b}\kern-0.8em\TeX}}}

\acmConference[CSE572, Data Mining, Progress Report]{}

\usepackage{subcaption}
\usepackage{graphicx}

\begin{document}

\title{Detecting and Refining HiRISE Image Patches Obscured by Atmospheric Dust}

\author{Kunal Kasodekar}
\email{kkasodek@asu.edu}
\affiliation{%
  \institution{SCAI}
  \institution{Arizona State University}
  \city{Tempe}
  \state{AZ}
  \country{USA}
}

\renewcommand{\shortauthors}{Kunal. Kasodekar}

\begin{abstract}

    HiRISE (High-Resolution Imaging Science Experiment) is a camera onboard the Mars Reconnaissance orbiter responsible for photographing vast areas of the Martian surface in unprecedented detail. It can capture millions of incredible closeup images in minutes. However, Mars suffers from frequent regional and local dust storms hampering this data-collection process, and pipeline, resulting in loss of effort and crucial flight time. Removing these images manually requires a large amount of manpower. I filter out these images obstructed by atmospheric dust automatically by using a Dust Image Classifier fine-tuned on Resnet-50 with an accuracy of 94.05\%. To further facilitate the seamless filtering of Images I design a prediction pipeline that classifies and stores these dusty patches.  I also denoise partially obstructed images using an Auto Encoder-based denoiser and Pix2Pix GAN with 0.75 and 0.99 SSIM Index respectively.
\end{abstract}



\maketitle

\section{Introduction}

A high-resolution imaging experiment (HiRISE for short) is a powerful camera on board the Mars Reconnaissance orbiter utilized for imaging the Martian surface up to a resolution of 3 cms/pixel. Its main objective is to study the Martian surface and investigate active surface processes and landscape evolutions as well as the potential for future human habitation/missions. It can capture images of the Martian surface that span large areas while also revealing details as minute as surface rocks. It can acquire large-scale image datasets (~28Gb) in less than 6 seconds. Thus managing, filtering and effectively storing a large set of these images is a necessity.

Mars suffers from periodic regional and local dust storms due to wind currents scooping up and moving the fine Martian dust to higher altitudes in the atmosphere. Thus, when HiRISE captures the Martian surface during these dusty conditions the images are redundant as the surface has been heavily obscured by dust storms. These storms occur sporadically and can hamper the data pipeline. Designated close-ups images required would not be successfully captured and would result in loss of flight time and interruptions to the data collection processes. 

I will make use of the dataset provided by Jet Propulsion Lab, named "HiRISE Image Patches Obscured by Atmospheric Dust." It is of utmost importance to reduce the processing time of the data pipeline, minimize manual efforts and optimize flight time. It is essential to discard dusty images at the source or at the very least effectively bifurcate and store them in a dataset for further processing. This processing can include denoising mission-critical data or vital images of rare Martian processes.

In order to achieve the aforementioned goals, I train various classifiers that can detect such dusty patches and create a pipeline using the best-performing model that can filter out these images from the dataset or at the source level. The dataset I am working with contains either fully obstructed or clean images. Hence, I plan to study the dust obscuring the images and create a new dataset with artificially imbibed noise that partially covers the image. I train an Auto-encoder and GAN model to denoise the images and recover the unobstructed Martian surface.

\begin{figure*}
\centering
\includegraphics{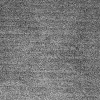}
\includegraphics{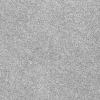}
\hspace{1.5 cm}
\includegraphics{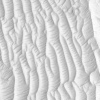}
\includegraphics{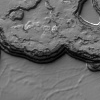}
\caption{Dusty Patches Obscured by Atmospheric Dust (left), Clear Images of the Martian Landscape (Right) [Resolution: 100x100]}
\end{figure*}

\section{Related Work}

There have been no studies on detecting, classifying or denoising Image Patches obscured by atmospheric dust. However, various studies have been done on denoising images and classifiers. 

Authors of Deep Mars \cite{Wagstaff_Lu_Stanboli_Grimes_Gowda_Padams_2018} do the multi-label classification of HiRISE orbital images using fine-tuning on HiRISENet and obtain a test accuracy of 94.5\%. The authors make use of transfer-learning on an Alexnet convolutional network backbone trained on Earth Images to do multi-label classification on Mars Images and create HiRISENet. With the help of this network, they develop a content-based search tool to find images of interest.

Images captured using optical telescopes like the Hubble Space Telescope inherently contain some perturbations and some additive noise. In order to maximize information gain and for further analysis, denoising is a mandatory preprocessing step. In the paper "Learning to Denoise Astronomical Images with U-nets" \cite{10.1093/mnras/staa3567} the authors propose a U-Net architecture called AstroNet which can recover around 95.9\% of the stars from such astronomical images.

The authors of "Generative Adversarial Networks Recover Features in Astrophysical Images of Galaxies beyond the Deconvolution Limit" \cite{10.1093/mnrasl/slx008} use GANs to reconstruct such perturbed galaxy images.

\section{Dataset}

I will make use of the dataset provided by Jet Propulsion Lab, named "HiRISE Image Patches Obscured by Atmospheric Dust" \cite{doran_gary_2019_3495068}. The dataset contains two sets of image patches each from the Experimental Data Record (i.e., Raw images without any corrections) and Reduced Data Record (i.e., corrected, and preprocessed images). Both these sets have 20,000 images each with training, validation, and test split of 10,000, 5,000 and 5,000 images each. No two splits have the same image. To estimate the inherent noise in the labels manually vetted labels are provided. I will not use K-folds cross-validation due to the inherently large size of the dataset. Both EDR and RDR subsets have an equal distribution of the classes i.e., the dataset is balanced. The image patches obtained from HiRISE are grayscale and 100 x 100 in dimensions. Currently, the plan is to experiment initially only with both RDR and later with EDR images due to its lack of comprehensive image correction and geometric processing (low-quality image data). The resolution of the RDR images is 100 x 100. Example images are provided in Figure 1.

\section{PROPOSED METHODOLOGY}

Initially, my goal was to train a classifier to detect dusty patches from clean Martian images. For classifying images as "Dusty" or "Not Dusty" Patches I do the following steps:

\subsection{Preprocessing}

The dataset contains a CSV file with image location, corresponding classes, and appropriate dataset split. I do the following steps:
\begin{itemize}
    \item I convert this file to a data frame, create image-to-class mappings and read all the images into python lists. Some initial data analysis and visualization are done to understand the characteristics of this dataset. 
    \item The images are normalized into the 0–1-pixel range and for the Resnet Model, I have another set of mean and standard deviation normalized images. These lists are converted to NumPy arrays and resized to sizes as per the classifiers used. For our PCA + SVM model, the data is reshaped/Flattened to (Batch Size, and image features) and for our CNN I resize it to 64 x 64 x 1.
    \item I also shuffle the data randomly, set seeds for multiple experiments and convert categorical labels, Dusty, and Not Dusty into a NumPy array with values 1 and 0 respectively. The Train-Val-Test split is 9817-5214-4969 images
\end{itemize}

\subsection{Classification}

Initially, I made use of Scikit-Learns PCA on our NumPy array with 10 components. I plot the explained variance ratio and pick our components with the max explained ratio i.e n=3 using our elbow point. Then I train an SVM with an RBF kernel and C=10000 on our data to predict our targets (Dusty/Not Dusty).

Then I created a simple convolutional neural network classifier in TensorFlow. The input shape is a 64x64x1 tensor followed by subsequent Convolution and Pooling layers with RELU activation. The final layers flatten the input, and the output is a 2-neuron dense layer with a SoftMax activation. The model is trained with Adam optimizer, Learning Rate = 3e-4 (Suggested by Andrej Karpathy) for 10 epochs with Sparse Categorical Cross entropy loss on 9817 images. The accuracy and loss per epoch are calculated. The predictions for the test images are analyzed to validate the model.

Finally, I experimented on a Resnet-50 Backbone in TensorFlow trained on Image Net in an attempt to achieve State of Art Results. I append the frozen Resnet Backbone with some dense and dropout layers after flattening the output (Following the pyramid model architecture). The final output layer has two neurons corresponding to the outputs and a Softmax activation. The images are resized (Upscaled) to a 224 x 224 image and stacked to convert the grayscale to RGB. The trainable linear probe is fine-tuned for 10 Epochs with the same aforementioned hyperparameters. The model predictions are tested on the test dataset to validate the accuracy.



\subsection{Image Noising}

It is crucial to model the noise/dusty patches recorded by the HiRISE Camera in order to train a model to denoise the images. The dusty patches first appear completely random, so this is what I did in my initial experiments is as follows:

I visually analyze the Dusty images to get an idea of the type of Noise present. Initially, I experiment with perturbing the images with salt-pepper noise using OpenCv2. To a predetermined group of pixels in each image, I added salt and pepper noise at random. Unfortunately, the pictures greatly differ from the actual dusty spots that were photographed. This is mostly due to two factors: first, the dusty patches are not limited to 255/0 sort of noisy pixels, i.e., noise is inherently grayscale; and second, even in the grayscale noise, there is a significant level of randomness. On the higher end of quantization, some images are formed entirely of noisy pixels, while others have light noisy pixels on the lower end. The images are nearly entirely made up of these noisy pixels, which appear to be a mix of light and dark grayscale pixels upon closer study. 

Further, I experimented with grayscale noisy values at the middle and higher end of the pixel spectrum (105-205 and 205-255) randomly on N points in the images. By manually adjusting the N values, grayscale noisy pixels, and proportion of pixel noise in the higher and lower spectrum I was able to get some visually similar images. 

To get an idea of the actual grayscale noise distribution I plot a histogram to find the peak values and ranges for these two levels of pixel quantization. I find that most of the pixels lie in the 70 to 220 value range with peaks at 90, 135 and 190.  

After experimenting with various methods to model the noise I finally settled for the following method:

To model the varying level of grayscale noise in the image, I design a function that adds noise randomly in between two ranges of quantization, higher and lower. In the lower level of quantization, I add noisy pixels from 70 to 145 and, for the higher range, I add noisy pixels between 145 to 220. The number of pixels I randomly add noise to will be decided based on the arguments of the function. 

\begin{figure*}
\centering
\includegraphics[scale=0.75]{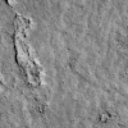}
\includegraphics[scale=0.75]{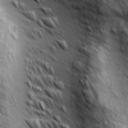}
\hspace{1.5 cm}
\includegraphics[scale=0.75]{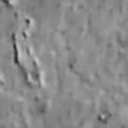}
\includegraphics[scale=0.75]{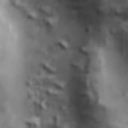}
\caption{Actual Clear Images (left), Denoised Cleaned Images using an Upscaled AutoEncoder Architecture (right) [Resolution: 128x128]}
\end{figure*}

\begin{figure*}
\centering
\includegraphics[scale=0.75]{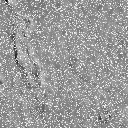}
\hspace{1.5 cm}
\includegraphics[scale=0.75]{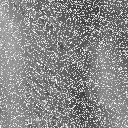}
\caption{Noisy Images for the Corresponding Denoised Images Above [Resolution: 128x128]}
\end{figure*}

\subsection{Denoising Models}

I have used an auto-encoder as a denoiser, as it can learn the latent representation of the images by compressing the inputs to a bottleneck layer and then reconstructing them. For a denoising auto-encoder, this property can be used to train the previous autoencoder to denoise noisy images at the input layer to their clean counterparts. However, these models are not extremely effective at denoising as the model tries to minimize the L2 loss across the image resulting in averaging out the outputs.

One could argue that a naive approach would be to train a CNN model for image-to-image translation with a Noisy input image and a clean denoised target image. Here the goal would be to minimize the pixel-to-pixel euclidean (L2) distance between the predicted and target image rather than having a high-level goal of producing an indistinguishable target image. Hence when such a kind of model is used it results in checkerboard effect/artificats at a large scale as the model tries to minimize its MSE loss across the whole image rather than focusing on minimizing actual semantic loss such that visual fidelity is not lost. Hence I make use of a conditional GAN called Pix2Pix that in addition to a latent vector for image generation, concatenates the corresponding denoised image. The generator in Pix2Pix makes use of a modified UNet with single double strided convolutions and skip connection at the decoder layers where the decoder layers are concatenated with the encoder layers. The generator along with its GAN loss has another L1(MAE) loss that minimizes the difference between real and generated target images. Generally, the discriminator's goal is to detect whether the image generated by the generator is real or fake. Here Pix2Pix makes use of PatchGAN (70x70) as the discriminator wherein it tries to classify every NxN as real or fake (Similar to a CNN) and then the results are averaged. These properties help the Pix2Pix model to not only fool the discriminator but also be as close as possible to the target in the L1 sense, faster in inference due to patching, better output quality due to skip connections in U-Net and produce an overall semantically similar image. Any blurry images produced by the generator will be quickly discarded by the Discriminator.

Initially, I train an auto-encoder with an input size of 100x100x1 pixels, 4 Encoding layers and a bottleneck/z layer of 25x25. I had experimented with adding more layers to reduce the bottleneck size but pushing for more encoding layers resulted in higher reconstruction loss. Due to the unconventional input size of 100x100 max-pooling odd-sized channels further in the network resulted in loss of features and spatial information. The clean, non-dusty images are preprocessed as above into 100x100x1 images and the Auto-Encoder is trained with the same input and output (Clean Not Dusty Images), with the same optimizer as above and for 100 epochs and a batch size of 64. Once trained the Auto-Encoder is retrained with input as Noised images generated above and clean images as the output. I experiment with various levels of noise and plot the results.

In order to alleviate some of these problems, I also upscale and downscale the images to 64x64 and 128x128 respectively and thus correspondingly change the model architecture by adding more layers and experimenting with a smaller bottleneck size of 8x8 (64), and 16x16 (256). After experimenting with all these architectures the upscaled deep architecture with a smaller bottleneck size of 16x16 (256) was working as the best denoiser (Architecture in Figure 4). A larger bottleneck size of z = 32x32 and shallow layers results in a more blurry reconstruction and the latent space doesn't capture the semantic understanding of the whole image. A highly compressed z=64 is not able to effectively reconstruct the denoised image from the latent space and suffers from blurring. 

Finally, I train a Pix2Pix model by using the Tensorflow tutorial \cite{Lamberta2022pix2pix} \cite{k_2022} and the Original Paper \cite{https://doi.org/10.48550/arxiv.1611.07004} as reference implementations. Initially, the conditional pair, (input, translated image) are prepared. For preprocessing the images from the training set are upscaled to a size of 256x256 as required by the network, normalized to a range of [-1,1] and augmented via multiple random image transformations. The model is compiled with a weighted sum of adversarial loss (BCE) and MAE loss with Adam optimizer, learning rate = 2e-4, and beta=0.5. The model is trained and evaluated after 10 Epochs.

\begin{figure}
\centering
\includegraphics[scale=0.5,height=325pt, width=150pt]{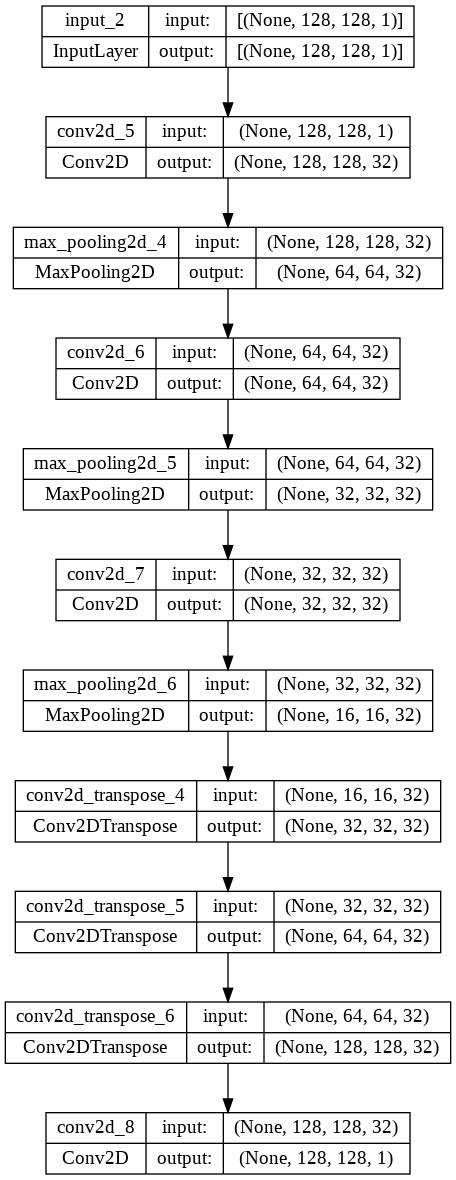}
\caption{Upscaled AutoEncoder Architecture}
\end{figure}

\section{Results}

\subsection{Classifier}

For predicting Dusty vs Not Dusty Images the SVM + PCA classifier achieves an accuracy of 61.96\% on the test data. I picked 3 Principal components using PCA as an input to SVM by finding the elbow point in the explained variance ratio graph. Averaging across multiple seeds has similar results. The Metrics and Confusion Matrix (Figure 5) for the model (test data) are below:\\

\begin{tabular}{ |c|c|c|c|c| } 
 \hline
     SVM+PCA  & precision & recall & f1-score & support \\ 
 \hline
 non-dusty & 0.60 & 0.68 & 0.64 & 2554 \\ 
 dusty & 0.65 & 0.56 & 0.60 & 2660 \\ [0.2cm]
 accuracy &  &  & 0.62 & 5214\\ 
 macro avg & 0.62 & 0.62 & 0.62 & 5214\\
 weighted avg & 0.62 & 0.62 & 0.62 & 5214\\
 \hline
\end{tabular}\\\\

\begin{tabular}{ |c|c| } 
 \hline
  Model & Accuracy\%(Test) \\ 
 \hline
 SVM + PCA &  61.96\\ 
 CNN & 89.74 \\ 
 Resnet50 (Finetuned) & 94.05 \\
 \hline
\end{tabular}\\

\noindent CNN achieves an accuracy of 93.16\% on the train data, 91.49\% on the validation data and 89.74\% on the test data with an F1 Score of 0.87(test). The Resnet50 model achieves the highest accuracy of 92.87\% on the train data, 95.03\% on the validation data and 94.05\% on the test data. The metrics for the Fine-Tuned Resnet50 model (test data) are as follows:\\

\begin{tabular}{ |c|c|c|c|c| } 
 \hline
    Resnet50   & precision & recall & f1-score & support \\ 
 \hline
 non-dusty & 0.95 & 0.93 & 0.94 & 2554 \\ 
 dusty & 0.93 & 0.95 & 0.94 & 2660 \\ [0.2cm]
 accuracy &  &  & 0.94 & 5214\\ 
 macro avg & 0.94 & 0.94 & 0.94 & 5214\\
 weighted avg & 0.94 & 0.94 & 0.94 & 5214\\
 \hline
\end{tabular}\\

\begin{figure}
\centering
\includegraphics[scale=0.5]{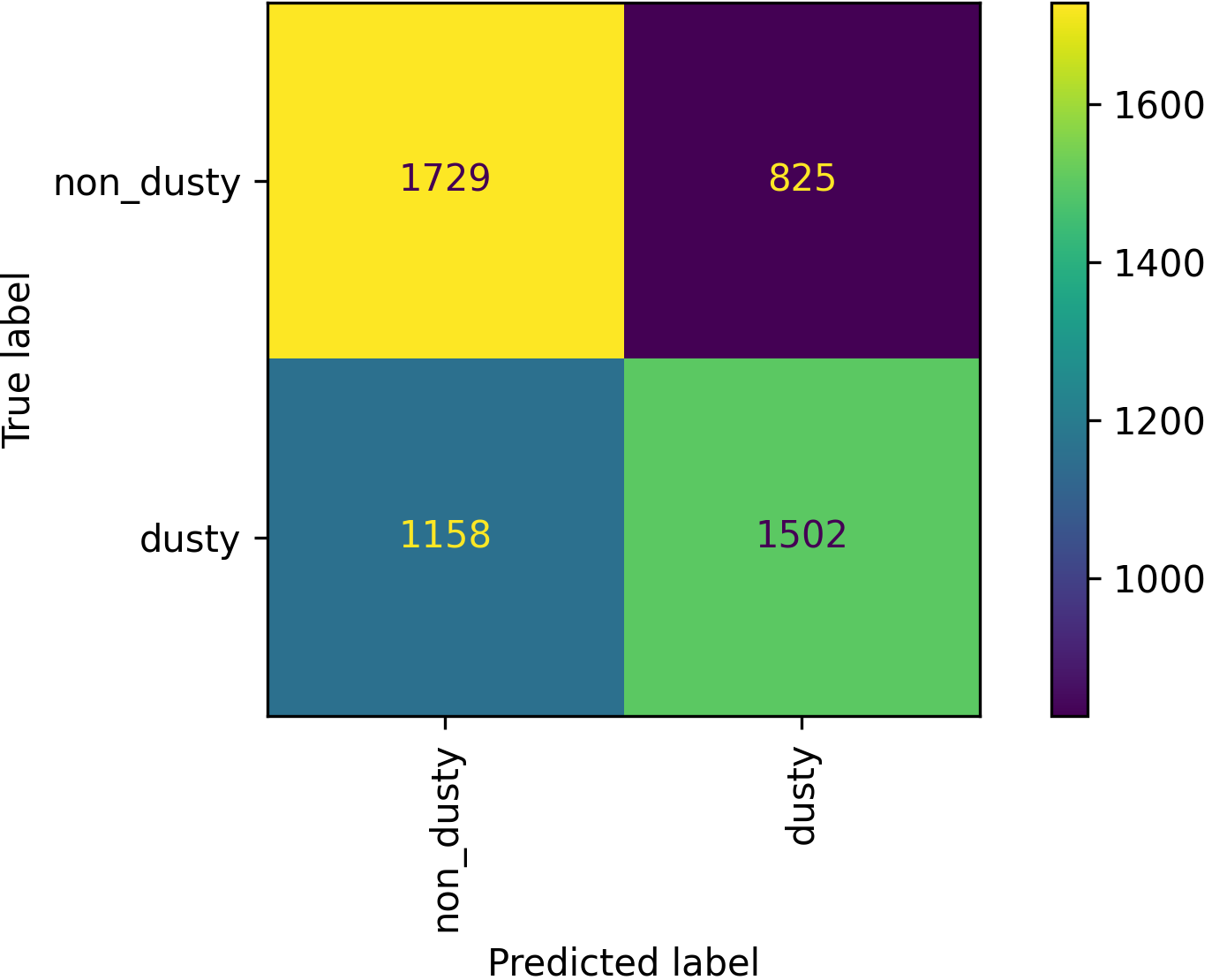}

\caption{Confusion Matrix}
\end{figure}

I also experimented with changing the number of layers in the linear probe, by increasing the dense layers and adding more dropout layers but the accuracy did not improve. I tried unfreezing the whole network and retraining it with a low learning rate of 1e-6 however the accuracy drastically reduced. For the CNN model, I try to use a LeNet-5-esque architecture with a receptive field of 18. The best model (Resnet50 in our case) is saved and used to create a pipeline that accepts an image folder, does preprocessing and automatically predicts and stores Dusty Images into an NPZ file. This filtering pipeline can be used to automatically filter out such dusty patches without manual intervention. We save all classified images into their respective arrays to an NPZ file for later use. Future uses can be denoising dusty Images and further analysis of classified images.

\subsection{Denoiser}

The initial Denoiser an Auto-Encoder with input size 100x100x1 with z=25x25 (625) trained to replicate input images achieves a loss of 0.6163 after 100 iterations. Once trained on the noisy data to clean the images the encoder achieves a loss of 0.6228. Based on the visual analysis the decoded cleaned images are somewhat clean of noise and slightly identical to the corresponding clean target images. However, there is massive blurring and loss of features in these denoised outputs (Figure 6). 

I further experiment with other Auto-Encoder Architectures designed based on our input image. I created two sets of images upscaled 128x128x1 images and downscaled 64x64x1 images and correspondingly design Auto-Encoders for these sizes. I add and reduce layers and change the size of the bottleneck layer.

Based on the validation loss and visual analysis upscaled images using an auto-encoder with a large bottleneck size and more convolutional layers are effective denoisers. The results can be inspected visually in Figures 2 and 3.

Then I train a Pix2Pix model and observe that the log-loss of the Discriminator is on average 0.350 and lower than log(2) i.e the discriminator performs better than the generator. The generator loss averages around 3.920. The denoised outputs from the Pix2Pix cgan lack artificats and look sharper and without loss of visual fidelity. Compared to the AutoEncoder the denoised images are visually cleaner. The results can be inspected visually in Figures 7 and 8.

Increasing the intensity of noisy pixels beyond 60-70\% of the image results in the auto-encoder model failing in denoising and producing blurry, pixelated outputs in many cases. However, the Pix2Pix model is able to effectively denoise the image even in this extreme case. On increasing the noise beyond 80\% of the pixels even the Pix2Pix model fails to restore the original clear image from the noisy image. The reason for this can be attributed to the fact that the input image is almost made of completely random noisy pixels and without sufficient clear pixels to determine the martian landscape features the model fails to generalize and denoise the image.

\subsection{Quality of Denoising}

After the denoising process, I obtained a somewhat clean denoised image. All my experiment results have been explained through human visual assessment, which relies on visual fidelity, perception and inspection. To use more concrete metrics to support my results. I calculate the following metrics on the test data:

\begin{itemize}
    \item \textbf{MAE}: It is the L1 pixel-to-pixel loss between the denoised and clear (Expected) image.
    \item \textbf{PSNR}: In the case of images it defines the ratio of the original data signal to the noisy error in our denoising process.
    \item \textbf{SSIM}: SSIM Index measures the similarity between two images and tells us how far an image is w.r.t to the expected/original image aligned with the human perceptual system.
    \item \textbf{Multiscale-SSIM}: It's the generalization of the SSIM Index at multiple image scales/resolutions.
\end{itemize}

It can see from the tables below that the Pix2Pix model performs better than the Upscaled AutoEncoder in terms of SSIM Score and MAE. It obtains a 0.99 SSIM score compared to 0.92 from the AutoEncoder thus indicating a higher visual similarity. Both the models fair almost equally for the other two metrics however, surprisingly AutoEncoder performs slightly better for these two metrics. Due to RAM Limitations, we compute the Metrics for the Pix2Pix image on only half the test images. The metrics for the denoising models can be viewed in the table below:\\


\begin{figure}
\centering
\includegraphics[scale=0.75]{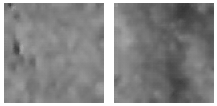}
\caption{Denoised Images with Base Encoder Architecture (Checkerboard Effect Observed) [Resolution: 100x100]}
\end{figure}

\begin{figure}
\centering
\includegraphics[scale=0.70]{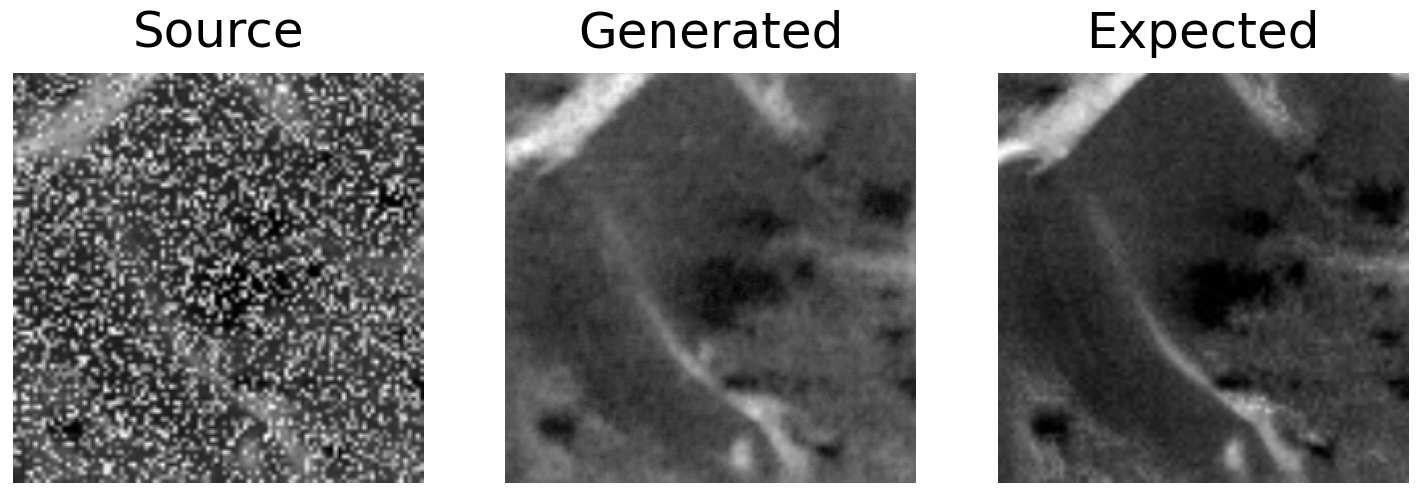}
\caption{Results from Pix2Pix. Images are highly perturbed with Random Noise[Resolution: 256x256]}
\end{figure}

\begin{figure}
\centering
\includegraphics[scale=0.70]{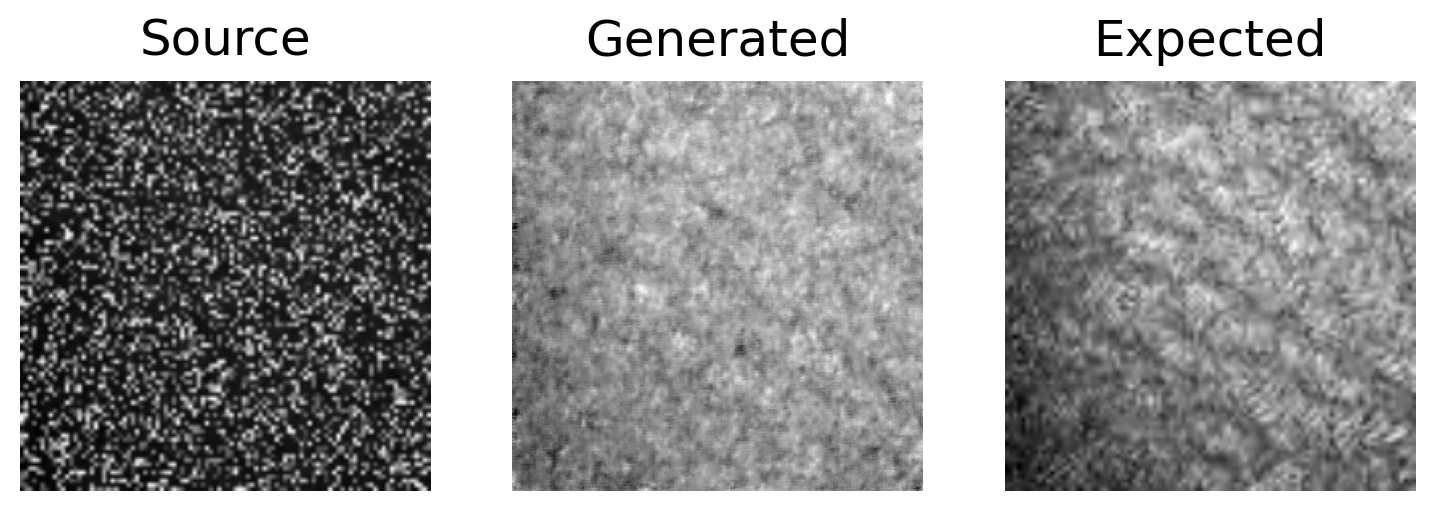}
\caption{More Results from Pix2Pix}
\end{figure}


\begin{tabular}{ |c|c|c| } 
 \hline
  Model & Metric & Value \\ 
 \hline
  & MAE (L1) &  0.022845278\\ 
  Upscaled  & SSIM (Mean) & 0.7573066 \\ 
  AutoEncoder & PSNR (Mean) & 31.497534 \\ 
   & Multiscale SSIM (Mean) & 0.9211203\\
 \hline
  & MAE (L1) &  0.023823868\\ 
  Pix2Pix & SSIM (Mean) & 0.9998866 \\ 
   & PSNR (Mean) & 30.1209 \\ 
   & Multiscale SSIM (Mean) & 0.88276297\\
   \hline
\end{tabular}


\section{Discussion}

Based on our observations from the aforementioned results I find that SVM+PCA with 3 principal components doesn’t work well and has a low F1 Score and accuracy. CNN works well as expected but doesn’t achieve state-of-the-art results. It doesn’t generalize well to the test set. Our Fine-Tuned Resnet50 classifier gives the best results, as expected. Resnet-50 is trained on Image Net and as a backbone will be a strong feature extractor. Freezing all the layers and replacing the final layer with a sequential model comprising a stack of linear layers creates an efficient classifier architecture. Fine-tuning this model results in creating a well-suited accurate classifier. Just creating and training is model is not enough. It is important to deploy this model for actual use, i.e classifying and storing dusty and clean patches in our use case. To achieve this goal a predictor pipeline is created that accepts an image folder loads it into an np array predicts the images and stores the results and images in an NPZ file.

For our denoiser, the initial denoiser architecture works well with its ability to denoise medium perturbed images. By upscaling the image and applying an Auto-Encoder with more layers and an optimal z value I get better results even in case of large perturbations. However, in extreme cases the decoded images are pixelated and lack textures due to the loss function's tendency to minimize pixel-to-pixel loss by averaging out the output witnessed by the low SSIM value and low MAE.

Finally, to overcome the issues from the previous denoising approaches I use the Pix2Pix model, an image-to-image translation model to denoise the image. The model effectively denoises images, without the repercussions faced by many of the other generative models witnessed by the extremely high SSIM value. It generates images, without artifacts that are sharp and with very few visual defects. As the Pix2Pix images are already highly upscaled naturally their MSSIM values will be slightly on the lower side. Although high levels of noise above the range of 80\% noisy pixels can affect its output this is possible to be circumvented by probably using more data, customizing the model architecture and training for more epochs.

\section{CONCLUSION}
I can successfully classify Dusty Patches obstructed by Atmospheric Dust from Non-Dusty clean Martian Patches. After experimenting with multiple models, I can conclude that my Fine-tuned Resnet50 with an accuracy of 94.05\% works the best. Further, I have developed a data pipeline to clean dusty images or filter them out and save them for further post-processing. One of these post-processing steps is denoising, and I was able to create a deep Auto-Encoder-based Denoiser that can denoise upscaled input images with decent resolution. I also train the Pix2Pix model to denoise highly perturbed images effectively with better denoising capabilities than all my other denoising models. In the case of highly perturbed images with too many noisy pixels, all the denoising models fail due to a lack of image features to work with for denoising. 

In the Future, I want to create and deploy an end-to-end classifier and denoising pipeline on the cloud and make it available for public use. I want to further experiment with EDR images and with designing a model that can learn and mimic the noise from the Dusty Patches and apply it to our clean images to generate training data. Finally, I want to create a denoiser that can filter out Dusty Patches at even 99\% noise.

\begin{acks}
This project, report and the research behind it would not have been possible without the support, insights and feedback from my supervisor Professor Hannah Kerner. The curriculum of CSE572 offered by her is extensive and covers all the essential concepts for Data Mining Research. Further, I would like to thank Liukis, Maria, Steven, et al.,  for open-sourcing the dataset, "HiRISE Image Patches Obscured by Atmospheric Dust", and for providing me with the opportunity to work on interesting and relevant real-world classification problems. Without this dataset, this project would not have been possible.
\end{acks}
\section*{Code}
\textbf{Github Link:} \url{https://github.com/gremlin97/Detecting-and-refining-HiRISE-Image-Patches-Obscured-by-Atmospheric-Dust}

\section*{SKILLS ACQUIRED}
I am thankful to Prof. Kerner for guiding me through this project, from proposal reviews to report revisions. Working on this project, I was able to acquire a wide array of skills, specifically those related to deep learning for Computer Vision and planetary remote sensing. I learned about the operation of the Mars Reconnaissance Orbiter (MRO), the HiRISE device onboard it, and the preprocessing considerations (EDR vs RDR) for planetary remote sensing devices. Furthermore, specifically for Computer Vision, I learned how to load, preprocess, and analyze planetary images for feature extraction, and then classify them using state-of-the-art image classifiers to detect noisy/denoised images. Additionally, I learned how to curate datasets with noisy images to train denoising models such as Auto Encoders and Pix2Pix GAN for denoising the images through the aforementioned models trained from scratch. I also learned how to properly report the results through classification reports and case-specific metrics such as MAE, PSNR, SSIM, and MultiScale-SSIM. Learning to train and work with the Pix2Pix GAN was a challenging and interesting aspect of this study. Finally, the most important lesson I learned was to identify relevant problems to apply data mining to and to find probable stakeholders for the same.

\section*{CONTRIBUTIONS}
This project is a solo endeavour conducted under the supervision of Prof. Hannah Kerner. I have performed all the experiments, written the code for them, and the report as well.

Dust storms frequently disrupt data collection, and subsequent manual removal of noisy images is inefficient. In order to automate this process, I created a Dust Image Classifier based on ResNet-50, which achieves 94.05\% accuracy. A prediction pipeline identifies and stores dusty patches, while an Auto Encoder-based denoiser and Pix2Pix GAN enhance partially obstructed images with SSIM Indexes of 0.75 and 0.99, respectively.

By acquiring the aforementioned skills, I have successfully classified patches from HiRISE MRO as Dusty or Non-Dusty using a fine-tuned ResNet-50 with an accuracy of 94.05\%. Furthermore, I developed a cascaded pipeline to segregate dusty patches and save them to disk for further filtration. For the filtered dusty patches, I created an AutoEncoder that upscales and denoises the dusty patches successfully. I also experimented with a Pix2Pix GAN for filtration, and in the case of low to mid noise, the model successfully denoises the dusty patches. However, in the case of highly perturbed images, the denoising is not successful.

\bibliographystyle{ACM-Reference-Format}
\bibliography{sample-base}
\end{document}